\documentclass[fleqn,10pt]{wlscirep}
\usepackage[utf8]{inputenc}
\usepackage[T1]{fontenc}
\usepackage{url}
\usepackage{longtable}
\usepackage{multirow}
\usepackage{enumitem}

\title{Privacy-Preserving Local Language Models for Longitudinal Data Retrieval in Chronic Dermatologic Disease: Implementation in Pemphigus Patients}

\author[1,*]{Abdurrahim Yilmaz, MS}
\author[2,*]{Ayşe Esra Koku Aksu, MD}
\author[2]{Duygu Yamen, MD}
\author[3]{Vefa Asli Erdemir, MD}
\author[3]{Mehmet Salih Gurel, MD}
\author[4]{Gulsum Gencoglan, MD}
\author[1]{Joram M. Posma, PhD}
\author[1]{Burak Temelkuran, PhD}

\affil[1]{Division of Systems Medicine, Department of Metabolism, Digestion and Reproduction, Imperial College London}
\affil[2]{Department of Dermatology and Venereology, Istanbul Research and Training Hospital}
\affil[3]{Department of Dermatology and Venereology, Istanbul Medeniyet University}
\affil[4]{Department of Dermatology and Venereology, Istanbul Medicana Atakoy Hospital}

\affil[*]{Corresponding Authors: \url{a.yilmaz23@imperial.ac.uk}, \url{ayseesra.kokuaksu@sbu.edu.tr}}

\begin{abstract}
  Chronic dermatologic diseases such as pemphigus require long-term follow-up, generating extensive longitudinal clinical documentation that is difficult to review comprehensively during routine visits and increasing clinician workload as well as the risk of missing critical historical information. We evaluated whether a locally deployed, privacy-preserving small language model (SLM) could retrieve structured clinical features and generate longitudinal summaries from long-term dermatology follow-up records. In this retrospective case series, thirty pemphigus patients contributed 541 visit notes that were aggregated into full longitudinal records (89,336 words); 56 clinically relevant features were annotated by two expert dermatologists. The locally deployed SLM (Qwen3 4B Thinking 2507) was queried with each complete record to retrieve 56 features and generate one final report summaries. Across 1,680 feature retrieval tasks, mean accuracy was 82.25\% ± 0.17\%. Dermatologists' ratings of AI-generated summaries were high for overall quality (8.23–8.47), clinical accuracy (7.93–8.20), and usefulness (8.47–8.50), with no significant inter-evaluator differences and an overall preference for AI summaries in 53.3\% of evaluations. These findings suggest that privacy-preserving, locally deployed SLMs can outperform medical experts and reliably generate clinically meaningful longitudinal summaries. SLMs may support clinical decision-making when integrated with appropriate oversight. \\\\\\

\end{abstract}

\begin{document}

\raggedbottom
\maketitle
\thispagestyle{empty}

\newpage

\section*{Introduction}

Chronic dermatologic diseases such as pemphigus necessitate prolonged, often lifelong follow-up, during which patients accumulate extensive clinical documentation spanning many years \cite{kasperkiewicz_pemphigus_2017, schmidt_pemphigus_2019}. These longitudinal records may include dozens of visits authored by different clinicians, each documenting disease activity, treatment adjustments, laboratory findings, laboratory results, and adverse events. In routine clinical practice, clinicians are expected to rapidly synthesize this historical information during follow-up visits to guide management decisions. However, comprehensive review of long and heterogeneous clinical narratives is frequently impractical in busy outpatient settings, particularly when patients are seen by different physicians over time. To manage this growing documentation burden, clinicians often rely on partial chart review or reuse of previous documentation through copying and pasting. While expedient, this practice introduces variability, reduces standardization, and increases the risk that critical historical details—such as prior exposure to rituximab, cumulative corticosteroid toxicity, or serious infections—are overlooked. In chronic autoimmune blistering diseases, where treatment decisions depend heavily on longitudinal disease behavior rather than isolated visits, incomplete recall of historical data may adversely affect patient safety and clinical outcomes.

Recent advances in artificial intelligence, particularly natural language processing (NLP) and language models (LMs), have demonstrated substantial potential for clinical summarization, information extraction, and clinical decision support \cite{van_veen_adapted_2024, wu_survey_2022, goldberg_primer_2016}. In dermatology, LM-based approaches have been explored for tasks ranging from disease classification to clinical documentation analysis, with growing interest in multimodal and domain-adapted models \cite{paganelli_natural_2024, azarfar_responsible_2025, yilmaz_resource-efficient_2025}. More broadly, LM-based systems have shown promise in generating clinical visit summaries, assisting with patient messaging, and supporting clinician communication workflows \cite{goodman-meza_natural_2022, bootsma-robroeks_ai-generated_2025}.

Despite these advances, most existing studies focus on short clinical texts such as discharge summaries, isolated visit notes, or synthetic tasks, and frequently rely on cloud-based inference as LMs requires expensive infrastructures \cite{marwaha_algorithmic_2025, wang_systematic_2020}. These approaches raise important concerns regarding patient privacy, data governance, and real-world deployability, particularly in healthcare systems with strict data protection requirements \cite{bear_dont_walk_scoping_2022, obika_safety_2024}. Longitudinal dermatology follow-up presents a particularly challenging testbed for clinical AI systems, as meaningful clinical insight depends on synthesizing information across years of care rather than single encounters. Resource-efficient and domain-specific AI models have recently been proposed to address such challenges in dermatology \cite{yilmaz_resource-efficient_2025}; however, systematic evaluations of local, privacy-preserving LMs on real-world longitudinal dermatology data remain limited.

In this study, we evaluate a fully local, privacy-preserving SLM-based system by using Qwen3 designed to retrieve structured clinical features and generate longitudinal summaries directly from complete multi-year dermatology follow-up records. Using pemphigus as a representative chronic dermatologic disease, we assess whether such a system can support automating longitudinal data retrieval, while maintaining high accuracy.

\section*{Methods}

\begin{figure}
  \centering
  \includegraphics[width=1\linewidth]{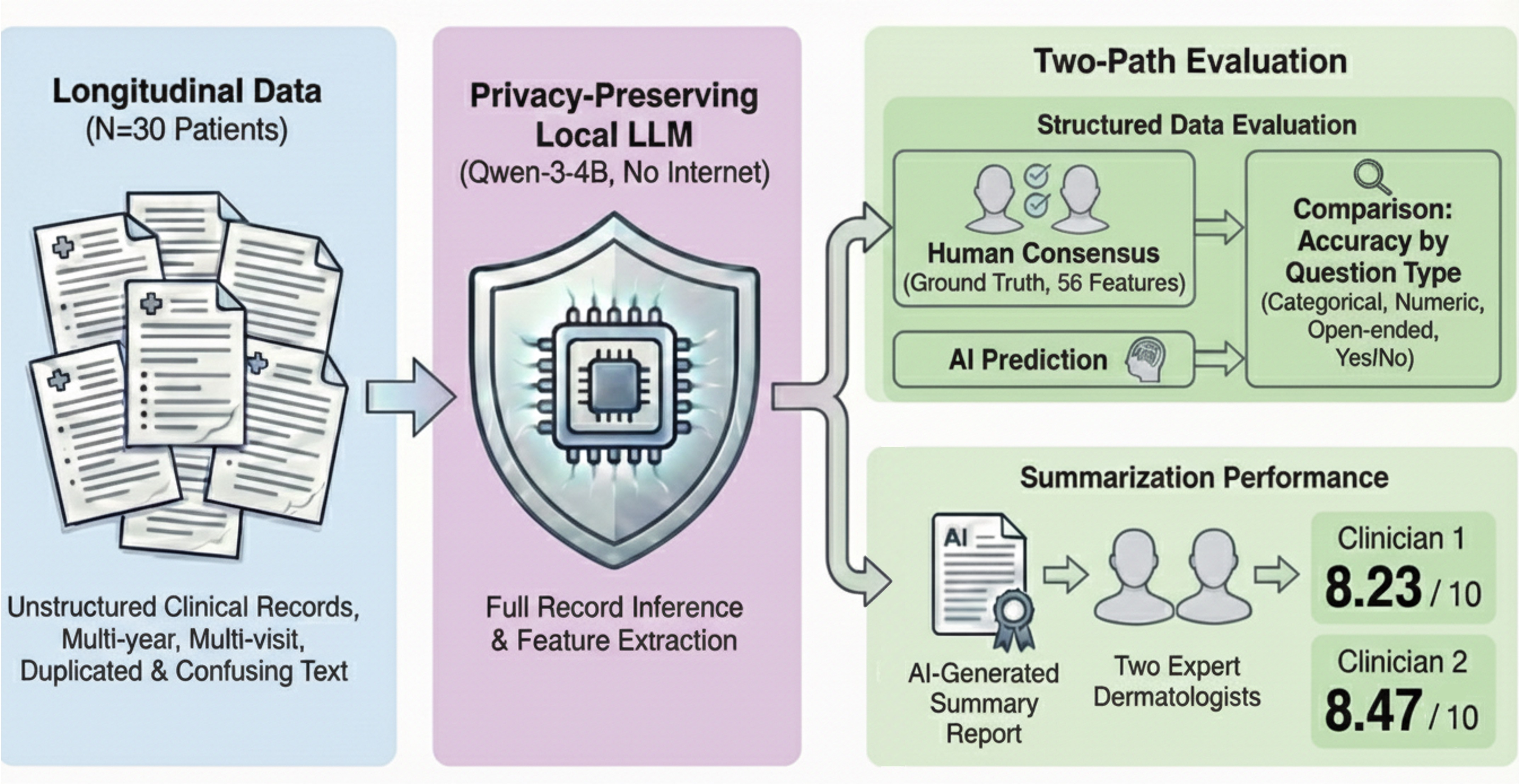}
  \caption{Overview of the study workflow: longitudinal pemphigus follow-up notes are aggregated into patient-level records and processed locally by a privacy-preserving SLM to retrieve structured clinical features and generate longitudinal summaries, which are then evaluated based on expert annotations.}
  \label{fig:ga}
\end{figure}

\subsection*{Study Design and Dataset}

The study was conducted in accordance with the Declaration of Helsinki and was approved by the Clinical Research Ethics Committee of Istanbul Research and Training Hospital (Approval No: 18, Date: 09 January 2026, Protocol No: 2011-KAEK-50). Given the retrospective design and use of de-identified clinical records, the requirement for informed consent was waived by the ethics committee. This retrospective case series included 30 patients diagnosed with pemphigus who underwent long-term dermatologic follow-up. All available visit notes for each patient were extracted from Istanbul Research and Training Hospital information system (DermaBase) and aggregated into a single chronological longitudinal record, preserving the original narrative text without manual summarization. The cohort comprised a total of 541 visit notes, with a mean of 18.03 ± 13.13\% visits per patient. The overview of the study workflow is presented at Figure \ref{fig:ga}.

\subsection*{Feature Schema and Ground Truth Annotation}
We defined a structured feature schema containing 56 features to capture clinically relevant longitudinal information from dermatology follow-up notes. These notes include diagnosis and phenotype, disease activity, laboratory findings (e.g., anti-desmoglein serology), systemic treatments (e.g., corticosteroids and steroid-sparing immunosuppressants), treatment dates and cycles (e.g., rituximab/IVIG), complications, and adverse events. The complete list of field names and their corresponding data types (categorical, numeric, or date) is provided in Table \ref{tab:labels}. In addition to structured fields, we evaluated generation of a clinician-facing final report (\textit{current\_status}) summarizing the longitudinal course. Ground truth annotations and summary report filling were performed by two expert dermatologists who had long-term experience managing the included patients (A.E.K.A. and D.Y.).

\begingroup
\renewcommand{\arraystretch}{1.05}
\begin{longtable}{p{0.72\linewidth}p{0.22\linewidth}}
  \caption{List of structured fields and their data types used for feature extraction.}\\
  \hline
  \textbf{Field name} & \textbf{Type} \\
  \hline
  \endfirsthead
  \hline
  \textbf{Field name} & \textbf{Type} \\
  \hline
  \endhead
  patient\_id & numeric \\
  visit\_count\_est & numeric \\
  pemphigus\_diagnosis & categorical \\
  cyclophosphamide\_ever & categorical \\
  rituximab\_ever & categorical \\
  rituximab\_cycles\_count & numeric \\
  rituximab\_first\_date & date \\
  rituximab\_last\_date & date \\
  rituximab\_response & categorical \\
  ivig\_ever & categorical \\
  ivig\_last\_date & date \\
  plasmapheresis\_ever & categorical \\
  plasmapheresis\_last\_date & date \\
  renal\_toxicity\_ever & categorical \\
  infusion\_reaction\_ever & categorical \\
  dose\_changes\_documented & categorical \\
  histology\_pemphigus\_compatible & categorical \\
  report\_id & categorical \\
  pemphigus\_subtype & categorical \\
  cytopenia\_ever & categorical \\
  report\_span\_end\_date & date \\
  phenotype\_mucosal & categorical \\
  serious\_infection\_ever & categorical \\
  diagnosis\_confidence & categorical \\
  anti\_dsg1\_positive\_at\_first\_diagnosis & numeric \\
  methotrexate\_ever & categorical \\
  steroid\_complication\_hypertension & categorical \\
  longitudinal\_course\_clear & categorical \\
  steroid\_complication\_myopathy & categorical \\
  dif\_pemphigus\_compatible & categorical \\
  anti\_dsg3\_positive\_at\_first\_diagnosis & numeric \\
  anti\_dsg3\_last\_value & numeric \\
  mmf\_ever & categorical \\
  steroid\_complication\_cataract & categorical \\
  anti\_dsg1\_last\_value & numeric \\
  report\_span\_start\_date & date \\
  steroid\_complication\_avascular\_necrosis & categorical \\
  azathioprine\_ever & categorical \\
  phenotype\_cutaneous & categorical \\
  liver\_toxicity\_ever & categorical \\
  steroid\_complication\_glaucoma & categorical \\
  steroid\_complications\_ever & categorical \\
  medication\_start\_stop\_dates\_present & categorical \\
  steroid\_complication\_hyperglycemia & categorical \\
  immunsupresant\_stop\_reason & categorical \\
  anti\_dsg3\_last\_date & date \\
  current\_systemic\_treatment & categorical \\
  anti\_dsg1\_last\_date & date \\
  current\_prednisone\_mg & numeric \\
  last\_flare\_date & date \\
  current\_disease\_activity & categorical \\
  steroid\_complication\_osteoporosis\_or\_osteopenia & categorical \\
  flare\_count\_est & numeric \\
  current\_steroid\_duration\_months & numeric \\
  total\_steroid\_duration\_months & numeric \\
  current\_status & text \\
  \hline
  \label{tab:labels}
\end{longtable}
\endgroup

\subsection*{Question Design and Inference Procedure}

We designed the inference procedure in two sequential steps to balance fair model evaluation with clinically meaningful interpretation. We initially evaluated the Qwen3 model using base prompts without additional domain-specific medical guidance to assess its intrinsic capability for longitudinal information extraction. Subsequently, for selected features that rely on standardized clinical definitions or implicit laboratory conventions, we incorporated minimal explanatory information directly into the main prompt to ensure fair interpretation (Appendix 1). This step was necessary because Qwen3 is not a medical-domain-trained model and may otherwise misinterpret closely related medical concepts.

For example, for the feature \texttt{steroid\_complication\_osteoporosis\_or\_osteopenia}, bone mineral density thresholds were explicitly provided (T-score below $-1$ indicating osteopenia; below $-2.5$ indicating osteoporosis). These clarifications were introduced solely to reduce ambiguity and did not involve advanced prompting strategies or multi-step reasoning. Each feature was paired with a feature-specific clinical question and an accompanying note specifying the expected answer format and interpretation rules (Supplementary Materials 2). For the final report (\textit{current\_status}), a bullet-based clinical template was provided to guide structured generation.

\subsection*{Model Deployment}

Inference was performed using the Qwen3 4B Thinking 2507 \cite{yang_qwen3_2025} model deployed locally via LM Studio (version 0.3.22) on a workstation equipped with an RTX 4090 Graphics Card with 24 GB VRAM and 96 GB RAM. Internet access was disabled, and all data remained local throughout the study. The model was accessed via OpenAI-compatible local endpoints using LM Studio's local hosting feature. Predictions were repeated 10 times using an in-house Python script (version 3.12.3). Inference parameters included a temperature of 0.3, a context length of 90,000 tokens, and a timeout-based stopping criterion of 100 seconds. Key study details are summarized in Table~\ref{tab:study_stats}.

\begin{table}[ht]
  \centering
  \caption{Summary of study and deployment statistics.}
  \label{tab:study_stats}
  \begin{tabular}{ll}
    \hline
    \textbf{Category} & \textbf{Value} \\
    \hline
    Patients & 30 \\
    Total visit notes & 541 \\
    Visits per patient & 18.03 (range: 1--40) \\
    Clinically relevant features & 56 \\
    Inference requests (single-pass) & 1,680 (56 $\times$ 30) \\
    Repeats & 10 \\
    Total model inferences & 16,800 \\
    Model & Qwen3 4B Thinking 2507 \\
    Deployment & LM Studio v0.3.22 \\
    Hardware & RTX 4090 (24 GB VRAM), 96 GB RAM \\
    Temperature & 0.3 \\
    Context length & 90,000 tokens \\
    Timeout & 100 s \\
    \hline
  \end{tabular}
\end{table}

\subsection*{Evaluation Metrics}

Structured features were evaluated using exact-match accuracy based on expert annotations. Open-ended features were evaluated using a two-tiered approach. First, automated text similarity was assessed using BLEU and ROUGE metrics. Second, clinical relevance and usefulness were evaluated through a double-blinded human assessment. Two expert dermatologists (each with $>20$ years of clinical experience) evaluated 30 AI-generated patient reports. For each patient, evaluators assessed both the original clinician-generated report and the AI-generated final report across four dimensions: (i) overall quality (10-point Likert scale), (ii) clinical accuracy (10-point Likert scale), (iii) clinical usefulness (10-point Likert scale), and (iv) overall preference (binary choice: clinician report vs AI report). A 10-point scale was selected to provide sufficient granularity for evaluating diverse, clinically meaningful error types that can occur in LM-generated reports. Reviewers were instructed to deduct points (typically 1--2 points per issue) for specific problems such as incorrect dates, incorrect clinical data, hallucinated statements, or clinically important missing information; smaller Likert ranges would have reduced sensitivity to these graded penalties and limited discrimination between near-correct and clearly incorrect outputs. Each evaluator completed two independent assessment sessions, resulting in 60 total evaluations (30 patients $\times$ 2 evaluators). Inter-rater reliability was assessed using the Intraclass Correlation Coefficient (ICC) for continuous measures. All results are presented as mean $\pm$ standard deviation (SD).

\section*{Results}

The AI system accepted each patient’s full longitudinal record as a single input, enabling feature retrieval from the entire follow-up history in one context. Specifically, the input consisted of the complete, chronologically aggregated free-text visit notes for a given patient. For each inference call, we queried the model for a single feature using the complete record, resulting in 1,680 feature-specific inference requests (56 features $\times$ 30 patients). To assess output stability, we repeated the full inference procedure ten times under identical conditions, yielding 16,800 total model inferences. Inference speed was also favorable, with a time-to-first-token of approximately 0.05 seconds and a generation speed of roughly 130 tokens per second (approximately 1 words equal 1.5 token). This performance allows rapid generation of longitudinal summaries, supporting real-time or near-real-time clinical use. Key performance metrics are summarized in Table~\ref{tab:results_metrics} for feature-based metrics and Table ~\ref{tab:mega_results} for statistical analysis.

\begin{table}[ht]
  \centering
  \caption{Summary of key performance results (mean $\pm$ SD across 10 repeated inference runs).}
  \label{tab:results_metrics}
  \begin{tabular}{ll}
    \hline
    \textbf{Metric} & \textbf{Value} \\
    \hline
    Overall accuracy & $82.25\% \pm 0.17\%$ \\
    Categorical feature accuracy & $86.34\% \pm 0.13\%$ \\
    Numerical feature accuracy & $70.61\% \pm 0.98\%$ \\
    Date-based feature accuracy & $80.56\% \pm 0.59\%$ \\
    Final report BLEU & $0.3602 \pm 0.0029$ \\
    Final report ROUGE-1 & $0.5195 \pm 0.0051$ \\
    Final report ROUGE-2 & $0.3873 \pm 0.0028$ \\
    Final report ROUGE-L & $0.4662 \pm 0.0055$ \\
    \hline
  \end{tabular}
\end{table}

Across all inference tasks, overall feature retrieval accuracy reached a mean of $82.25\%$ ($\pm 0.17\%$; min $82.00\%$; max $82.67\%$). Feature-level performance varied by data type. Categorical features achieved a mean exact-match accuracy of $86.34\%$ ($\pm 0.13\%$; min $86.10\%$, max $86.57\%$), while numerical features reached $70.61\%$ ($\pm 0.98\%$; min $70.00\%$, max $73.33\%$) and date-based features $80.56\%$ ($\pm 0.59\%$; min $78.89\%$, max $80.74\%$). Errors in numerical and date fields most commonly involved imprecise value extraction or partial temporal ambiguity rather than gross misclassification. For the open-ended final report, automated text similarity evaluation yielded a mean BLEU score of $0.3602$ ($\pm 0.0029$), a ROUGE-1 score of $0.5195$ ($\pm 0.0051$), a ROUGE-2 score of $0.3873$ ($\pm 0.0028$), and a ROUGE-L score of $0.4662$ ($\pm 0.0055$). Ground-truth reports had an average length of 758.9 characters (105.1 words), whereas model-generated reports averaged 1580.8 characters (217.7 words), corresponding to a mean increase of +821.9 characters (+108.3\%) and +112.6 words (+107.1\%).

\begin{table*}[ht]
  \centering
  \caption{Summary of Report Characteristics, Expert Evaluation, Preference, and Reliability Analyses}
  \renewcommand{\arraystretch}{1}
  \begin{tabular}{llcc}
    \hline
    \textbf{Category} & \textbf{Metric} & \textbf{Result} \\
    \hline
    \multirow{4}{*}{Report Length}
    & Ground-truth length & 758.9 characters (105.1 words)  \\
    & Model-generated length & 1580.8 characters (217.7 words)  \\
    & Absolute increase & +821.9 characters, +112.6 words  \\
    & Relative increase & +108.3\% characters, +107.1\% words  \\
    \hline
    \multirow{3}{*}{Expert Evaluation}
    & Overall quality & 8.23 vs 8.47 (Eval 1 vs Eval 2)  \\
    & Clinical accuracy & 7.93 vs 8.20 (Eval 1 vs Eval 2)  \\
    & Clinical usefulness & 8.50 vs 8.47 (Eval 1 vs Eval 2)  \\
    \hline
    \multirow{4}{*}{Preference Analysis}
    & Overall preference & 32/60 - AI 53.3\% vs Clinician 46.7\%  \\
    & Evaluation 1 preference & AI 43.3\% vs Clinician 56.7\% \\
    & Evaluation 2 preference & AI 63.3\% vs Clinician 36.7\%  \\
    & Total evaluations & 60  \\
    \hline
    \multirow{3}{*}{Reliability \& Correlation}
    & Inter-session reliability (ICC) & 0.648--0.723  \\
    & Metric correlations (Pearson $r$) & 0.92--0.97  \\
    & Preference vs quality correlation & 0.70--0.81  \\
    \hline
  \end{tabular}
  \label{tab:mega_results}
\end{table*}

In addition to automated metrics, the model-generated final report was evaluated using a double-blinded 10-point Likert-scale assessment by two expert dermatologists with $>20$ years of clinical experience. A total of 60 evaluations were completed. Quality scores were high across all metrics. Mean scores were: overall quality 8.23 (Evaluation 1) vs 8.47 (Evaluation 2; $p = 0.147$), clinical accuracy 7.93 (Evaluation 1) vs 8.20 (Evaluation 2; $p = 0.133$), and clinical usefulness 8.50 (Evaluation 1) vs 8.47 (Evaluation 2; $p = 0.851$). Paired $t$-tests revealed no statistically significant differences between evaluation sessions for any metric (all $p > 0.05$).

In preference analysis (60 total evaluations), AI-generated reports were preferred in 53.3\% of evaluations (32/60), compared to 46.7\% (28/60) for clinician-generated reports. Preference patterns differed between evaluation sessions: Evaluation 1 favored clinician reports (Ground Truth: 56.7\% vs AI: 43.3\%), whereas Evaluation 2 favored AI reports (Ground Truth: 36.7\% vs AI: 63.3\%). Reliability analysis showed moderate agreement between evaluation sessions for continuous metrics (ICC = 0.648--0.723, all $p < 0.001$). Correlation analysis demonstrated strong positive correlations between quality metrics within each evaluation (Pearson $r = 0.92$--0.97, all $p < 0.001$). Preference was positively associated with quality scores ($r = 0.70$--0.81), suggesting that reports rated more highly were more often selected as preferred.

\section*{Discussion}

This study demonstrates that a fully local, privacy-preserving SLM can retrieve clinically meaningful longitudinal information from complex dermatology follow-up records. By automating longitudinal data retrieval and generating clinician-facing reports, such systems have the potential to reduce clinician workload during routine follow-up visits \cite{dymek_building_2021, gandhi_how_2023}. Importantly, our findings emphasize the role of LMs primarily as report-generation and summarization tools, rather than autonomous clinical decision-makers. The low standard deviations observed for type-wise structured extraction (categorical, numeric, and date fields) indicate high reproducibility across repeated inference runs under fixed decoding settings, even when mean performance differed by field type.

\begin{figure}
  \centering
  \includegraphics[width=1\linewidth]{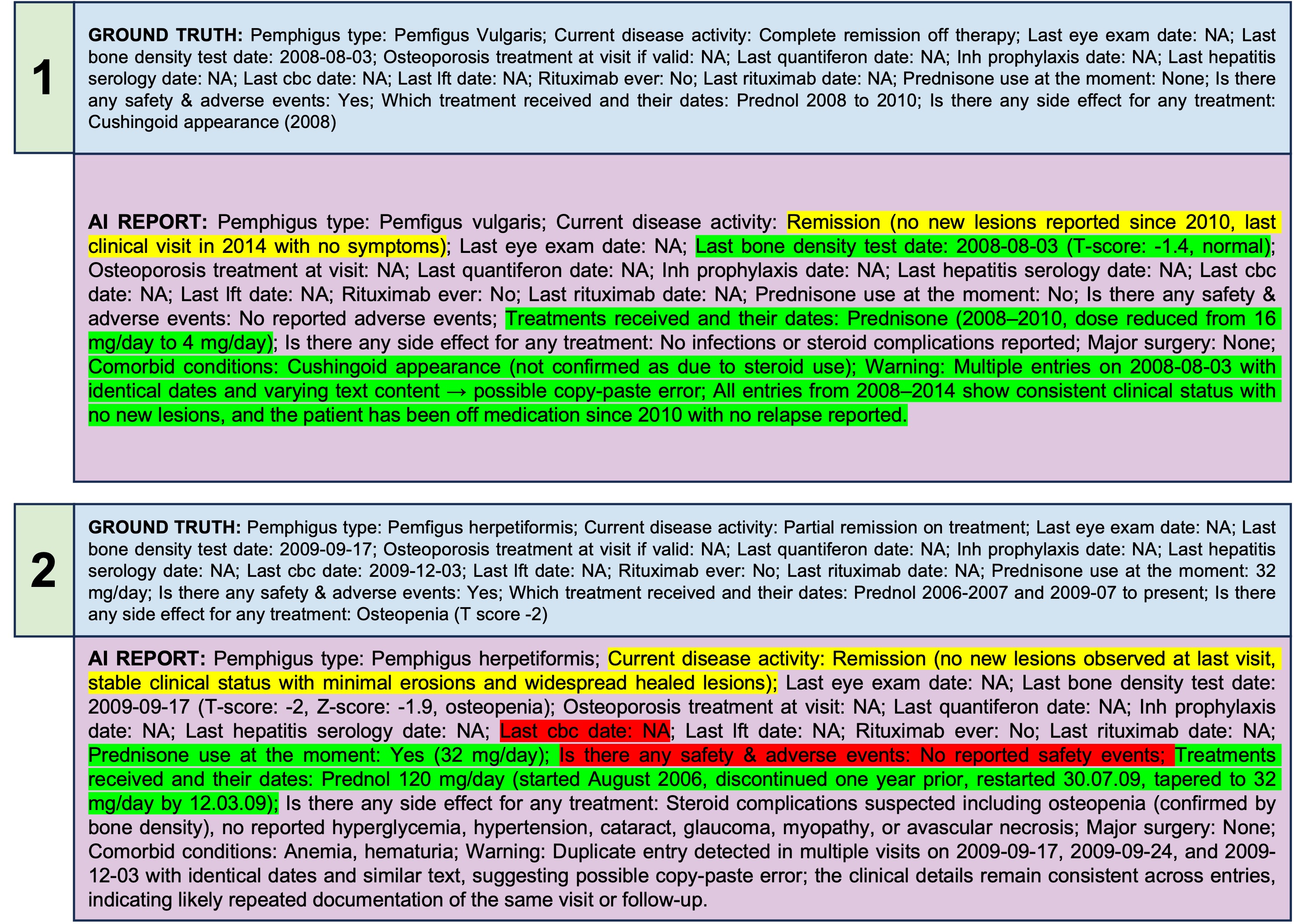}
  \caption{Qualitative comparison between clinician-written ground truth reports and AI-generated reports: Each example consists of two parts, with the clinician-written ground truth report shown above and the AI-generated response shown below. The figure contains text-only excerpts annotated for qualitative analysis. Correct or clinically useful information is highlighted in green, explanatory content that does not align with the instruction is highlighted in yellow, and clear errors are highlighted in red. (1) Example of a generally good AI report: although disease activity is not detected correctly, the explanation is helpful, treatments and comorbidities are well captured, and the warning statements are useful for clinician review. (2) Example of a poor AI report: despite clear explanations and reasonable coverage of treatments and comorbidities, disease activity is not detected correctly, limiting overall report quality. Both AI-generated reports are longer than the ground truth reports, as the AI model provides additional explanatory commentary. Clinicians prepared the ground truth reports for automated metric calculation (BLEU and ROUGE), with primary comparisons conducted based on the original reports. This figure is included solely to support qualitative visualization and interpretation of model behavior.}
  \label{fig:example}
\end{figure}

From a clinical perspective, the ability to rapidly surface key longitudinal information (such as prior biologic exposure, cumulative corticosteroid burden, flare frequency, and serious adverse events) may improve clinical decision-making and reduce cognitive load. This is particularly relevant in longitudinal dermatology care, where patients are often managed by multiple clinicians over extended periods.

A central finding of this study is that meaningful clinical benefit arises primarily from automated report generation for clinicians, rather than perfect extraction of every individual feature. The evaluated Qwen3 ``thinking'' model appears better suited to narrative synthesis (final-report writing) than strict structured extraction: while it missed some details during feature-level retrieval, it could often integrate the same information more coherently when generating the final report \cite{center_for_ai_safety_benchmark_2026}. This behavior also highlights that different LMs have different capabilities for different clinical tasks \cite{sellergren_medgemma_2025}. No single small model can optimally handle all clinical NLP tasks, model selection should be driven by the intended workflow. In this study, our primary aim was longitudinal report generation, while also systematically assessing feature-based clinical data retrieval. In contrast, the Qwen3 4B “thinking” model demonstrated superior report-generation performance, likely due to its stronger reasoning capabilities and ability to integrate long-context information.

Failure analysis suggested that model performance improved when prompts included sufficient task-specific detail and explicit decision rules. In some cases, we labeled the phenotype as ``no'' under our study definition because only a suspicion was documented, but the model initially answered ``yes'' and justified this choice during its internal ``thinking'' process. When the prompt was refined to explicitly encode the study definition and instruct the model to treat ``suspicion'' as insufficient evidence, the model could be aligned to produce the correct output, effectively bypassing its intrinsic default assumptions. These observations suggest that a subset of feature-level errors reflect prompt ambiguity rather than irrecoverable model limitations, and that definition-aware prompting or structured inputs may reduce clinically relevant mistakes in real-world deployments.

Crucially, this study was designed to mimic daily clinical practice as closely as possible. Clinician notes were provided to the model exactly as stored in the hospital information system (DermaBase), without preprocessing, cleaning, or human intervention. This ``in-the-wild'' evaluation allowed us to directly observe common challenges faced by LM-based clinical systems, most notably hallucination and misinformation generation. In contrast to many clinical information-extraction studies that evaluate LMs on pre-structured fields exported from hospital systems, our evaluation used raw longitudinal narrative documentation, which better reflects real-world clinical variability and incomplete recording practices. For instance, the model occasionally confused treatments, particularly regarding systemic corticosteroid (prednol) use, likely due to duplicated or inconsistently documented entries across visits. The model also made incorrect causal assumptions, such as interpreting hepatosteatosis as evidence of drug-induced liver toxicity and labeling the test date as NA when it occurred in the distant past (Figure \ref{fig:example}). While many of these issues could be addressed through more detailed prompts or standardized documentation, their occurrence highlights real-world risks when deploying LMs in unstructured clinical environments \cite{shah_accuracy_2024, du_testing_2024, kim_medical_2025}.

Feedback collected during the Likert-scale evaluation provided additional insight into the perceived quality of AI-generated reports. Most clinician comments focused on report structure, formatting, or missing preferred data elements (e.g., inclusion of specific laboratory values or treatment timelines). Notably, these issues were largely related to presentation rather than factual correctness and could likely be addressed with relatively simple prompt engineering or template-based constraints.



Our limitations should be acknowledged. First, pemphigus is a rare disease, and larger multi-center datasets will be required to validate these findings and improve the precision of performance estimates. Second, this was a single-center study, which may limit generalizability to institutions with different documentation styles or clinical workflows.

In conclusion, local LMs represent a promising, privacy-preserving approach for longitudinal data retrieval and report generation in dermatology. Future work should focus on prospective evaluation of clinician workload reduction, expansion to other chronic diseases, and integration into real-world clinical workflows. For future clinical decision-making studies, report-generation quality may be further improved by providing richer, better-grounded inputs (e.g., integrating PDF documents and structured summaries) and by combining local models with integrations such as retrieval-augmented generation (RAG) pipelines to reduce missing context and improve factuality. If local models can achieve consistently reliable longitudinal information synthesis across diverse settings, they may serve as a foundational component of privacy-preserving clinical decision support systems in future studies.

\section*{Acknowledgements}

Conflict of Interest Disclosure: None Declared. \\

\noindent Abdurrahim Yilmaz has been funded by the President's PhD Scholarships of Imperial College London. \\


\bibliography{skin}

@article{shah_accuracy_2024,
	title = {Accuracy, {Consistency}, and {Hallucination} of {Large} {Language} {Models} {When} {Analyzing} {Unstructured} {Clinical} {Notes} in {Electronic} {Medical} {Records}},
	volume = {7},
	issn = {2574-3805},
	url = {https://jamanetwork.com/journals/jamanetworkopen/fullarticle/2822301},
	doi = {10.1001/jamanetworkopen.2024.25953},
	language = {en},
	number = {8},
	urldate = {2026-02-08},
	journal = {JAMA Network Open},
	author = {Shah, Savyasachi V.},
	month = aug,
	year = {2024},
	pages = {e2425953},
}

@misc{du_testing_2024,
	title = {Testing and {Evaluation} of {Generative} {Large} {Language} {Models} in {Electronic} {Health} {Record} {Applications}: {A} {Systematic} {Review}},
	copyright = {http://creativecommons.org/licenses/by/4.0/},
	shorttitle = {Testing and {Evaluation} of {Generative} {Large} {Language} {Models} in {Electronic} {Health} {Record} {Applications}},
	url = {http://medrxiv.org/lookup/doi/10.1101/2024.08.11.24311828},
	doi = {10.1101/2024.08.11.24311828},
	language = {en},
	urldate = {2026-02-08},
	publisher = {Health Informatics},
	author = {Du, Xinsong and Zhou, Zhengyang and Wang, Yifei and Chuang, Ya-Wen and Li, Yiming and Yang, Richard and Zhang, Wenyu and Wang, Xinyi and Chen, Xinyu and Guan, Hao and Lian, John and Hong, Pengyu and Bates, David W. and Zhou, Li},
	month = aug,
	year = {2024},
}

@misc{kim_medical_2025,
	title = {Medical {Hallucinations} in {Foundation} {Models} and {Their} {Impact} on {Healthcare}},
	url = {http://arxiv.org/abs/2503.05777},
	doi = {10.48550/arXiv.2503.05777},
	urldate = {2026-02-08},
	publisher = {arXiv},
	author = {Kim, Yubin and Jeong, Hyewon and Chen, Shan and Li, Shuyue Stella and Park, Chanwoo and Lu, Mingyu and Alhamoud, Kumail and Mun, Jimin and Grau, Cristina and Jung, Minseok and Gameiro, Rodrigo and Fan, Lizhou and Park, Eugene and Lin, Tristan and Yoon, Joonsik and Yoon, Wonjin and Sap, Maarten and Tsvetkov, Yulia and Liang, Paul and Xu, Xuhai and Liu, Xin and Park, Chunjong and Lee, Hyeonhoon and Park, Hae Won and McDuff, Daniel and Tulebaev, Samir and Breazeal, Cynthia},
	month = nov,
	year = {2025},
	note = {arXiv:2503.05777 [cs]},
}

@article{center_for_ai_safety_benchmark_2026,
	title = {A benchmark of expert-level academic questions to assess {AI} capabilities},
	volume = {649},
	issn = {0028-0836, 1476-4687},
	url = {https://www.nature.com/articles/s41586-025-09962-4},
	doi = {10.1038/s41586-025-09962-4},
	language = {en},
	number = {8099},
	urldate = {2026-02-08},
	journal = {Nature},
	author = {{Center for AI Safety} and Phan, Long and Gatti, Alice and Li, Nathaniel and Khoja, Adam and Kim, Ryan and Ren, Richard and Hausenloy, Jason and Zhang, Oliver and Mazeika, Mantas and Hendrycks, Dan and {Scale AI} and Han, Ziwen and Hu, Josephina and Zhang, Hugh and Zhang, Chen Bo Calvin and Shaaban, Mohamed and Ling, John and Shi, Sean and Choi, Michael and Agrawal, Anish and Chopra, Arnav and Nattanmai, Aakaash and McKellips, Gordon and Cheraku, Anish and Suhail, Asim and Luo, Ethan and Deng, Marvin and Luo, Jason and Zhang, Ashley and Jindel, Kavin and Paek, Jay and Halevy, Kasper and Baranov, Allen and Liu, Michael and Avadhanam, Advaith and Zhang, David and Cheng, Vincent and Ma, Brad and Fu, Evan and Do, Liam and Lass, Joshua and Yang, Hubert and Sunkari, Surya and Bharath, Vishruth and Ai, Violet and Leung, James and Agrawal, Rishit and Zhou, Alan and Chen, Kevin and Kalpathi, Tejas and Xu, Ziqi and Wang, Gavin and Xiao, Tyler and Maung, Erik and Lee, Sam and Yang, Ryan and Yue, Roy and Zhao, Ben and Yoon, Julia and Sun, Xiangwan and Singh, Aryan and Peng, Clark and Osbey, Tyler and Wang, Taozhi and Echeazu, Daryl and Wu, Timothy and Patel, Spandan and Kulkarni, Vidhi and Sundarapandiyan, Vijaykaarti and Le, Andrew and Nasim, Zafir and Yalam, Srikar and Kasamsetty, Ritesh and Samal, Soham and Sun, David and Shah, Nihar and Saha, Abhijeet and Zhang, Alex and Nguyen, Leon and Nagumalli, Laasya and Wang, Kaixin and Wu, Aidan and Telluri, Anwith and Yue, Summer and Wang, Alexandr and {HLE Contributors Consortium} and Dodonov, Dmitry and Nguyen, Tung and Lee, Jaeho and Anderson, Daron and Doroshenko, Mikhail and Stokes, Alun Cennyth and Mahmood, Mobeen and Pokutnyi, Oleksandr and Iskra, Oleg and Wang, Jessica P. and Levin, John-Clark and Kazakov, Mstyslav and Feng, Fiona and Feng, Steven Y. and Zhao, Haoran and Yu, Michael and Gangal, Varun and Zou, Chelsea and Wang, Zihan and Popov, Serguei and Gerbicz, Robert and Galgon, Geoff and Schmitt, Johannes and Yeadon, Will and Lee, Yongki and Sauers, Scott and Sanchez, Alvaro and Giska, Fabian and Roth, Marc and Riis, Søren and Utpala, Saiteja and Burns, Noah and Goshu, Gashaw M. and Naiya, Mohinder Maheshbhai and Agu, Chidozie and Giboney, Zachary and Cheatom, Antrell and Fournier-Facio, Francesco and Crowson, Sarah-Jane and Finke, Lennart and Cheng, Zerui and Zampese, Jennifer and Hoerr, Ryan G. and Nandor, Mark and Park, Hyunwoo and Gehrunger, Tim and Cai, Jiaqi and McCarty, Ben and Garretson, Alexis C. and Taylor, Edwin and Sileo, Damien and Ren, Qiuyu and Qazi, Usman and Li, Lianghui and Nam, Jungbae and Wydallis, John B. and Arkhipov, Pavel and Shi, Jack Wei Lun and Bacho, Aras and Willcocks, Chris G. and Cao, Hangrui and Motwani, Sumeet and De Oliveira Santos, Emily and Veith, Johannes and Vendrow, Edward and Cojoc, Doru and Zenitani, Kengo and Robinson, Joshua and Tang, Longke and Li, Yuqi and Vendrow, Joshua and Fraga, Natanael Wildner and Kuchkin, Vladyslav and Maksimov, Andrey Pupasov and Marion, Pierre and Efremov, Denis and Lynch, Jayson and Liang, Kaiqu and Mikov, Aleksandar and Gritsevskiy, Andrew and Guillod, Julien and Demir, Gözdenur and Martinez, Dakotah and Pageler, Ben and Zhou, Kevin and Soori, Saeed and Press, Ori and Tang, Henry and Rissone, Paolo and Green, Sean R. and Brüssel, Lina and Twayana, Moon and Dieuleveut, Aymeric and Imperial, Joseph Marvin and Prabhu, Ameya and Yang, Jinzhou and Crispino, Nick and Rao, Arun and Zvonkine, Dimitri and Loiseau, Gabriel and Kalinin, Mikhail and Lukas, Marco and Manolescu, Ciprian and Stambaugh, Nate and Mishra, Subrata and Hogg, Tad and Bosio, Carlo and Coppola, Brian P. and Salazar, Julian and Jin, Jaehyeok and Sayous, Rafael and Ivanov, Stefan and Schwaller, Philippe and Senthilkumar, Shaipranesh and Bran, Andres M. and Algaba, Andres and Van Den Houte, Kelsey and Van Der Sypt, Lynn and Verbeken, Brecht and Noever, David and Kopylov, Alexei and Myklebust, Benjamin and Li, Bikun and Schut, Lisa and Zheltonozhskii, Evgenii and Yuan, Qiaochu and Lim, Derek and Stanley, Richard and Yang, Tong and Maar, John and Wykowski, Julian and Oller, Mart and Sahu, Anmol and Ardito, Cesare Giulio and Hu, Yuzheng and Kamdoum, Ariel Ghislain Kemogne and Jin, Alvin and Vilchis, Tobias Garcia and Zu, Yuexuan and Lackner, Martin and Koppel, James and Sun, Gongbo and Antonenko, Daniil S. and Chern, Steffi and Zhao, Bingchen and Arsene, Pierrot and Cavanagh, Joseph M. and Li, Daofeng and Shen, Jiawei and Crisostomi, Donato and Zhang, Wenjin and Dehghan, Ali and Ivanov, Sergey and Perrella, David and Kaparov, Nurdin and Zang, Allen and Sucholutsky, Ilia and Kharlamova, Arina and Orel, Daniil and Poritski, Vladislav and Ben-David, Shalev and Berger, Zachary and Whitfill, Parker and Foster, Michael and Munro, Daniel and Ho, Linh and Sivarajan, Shankar and Hava, Dan Bar and Kuchkin, Aleksey and Holmes, David and Rodriguez-Romero, Alexandra and Sommerhage, Frank and Zhang, Anji and Moat, Richard and Schneider, Keith and Kazibwe, Zakayo and Clarke, Don and Kim, Dae Hyun and Dias, Felipe Meneguitti and Fish, Sara and Elser, Veit and Kreiman, Tobias and Vilchis, Victor Efren Guadarrama and Klose, Immo and Anantheswaran, Ujjwala and Zweiger, Adam and Rawal, Kaivalya and Li, Jeffery and Nguyen, Jeremy and Daans, Nicolas and Heidinger, Haline and Radionov, Maksim and Rozhoň, Václav and Ginis, Vincent and Stump, Christian and Cohen, Niv and Poświata, Rafał and Tkadlec, Josef and Goldfarb, Alan and Wang, Chenguang and Padlewski, Piotr and Barzowski, Stanislaw and Montgomery, Kyle and Stendall, Ryan and Tucker-Foltz, Jamie and Stade, Jack and Rogers, T. Ryan and Goertzen, Tom and Grabb, Declan and Shukla, Abhishek and Givré, Alan and Ambay, John Arnold and Sen, Archan and Aziz, Muhammad Fayez and Inlow, Mark H. and He, Hao and Zhang, Ling and Kaddar, Younesse and Ängquist, Ivar and Chen, Yanxu and Wang, Harrison K. and Ramakrishnan, Kalyan and Thornley, Elliott and Terpin, Antonio and Schoelkopf, Hailey and Zheng, Eric and Carmi, Avishy and Brown, Ethan D. L. and Zhu, Kelin and Bartolo, Max and Wheeler, Richard and Stehberger, Martin and Bradshaw, Peter and Heimonen, Jp and Sridhar, Kaustubh and Akov, Ido and Sandlin, Jennifer and Makarychev, Yury and Tam, Joanna and Hoang, Hieu and Cunningham, David M. and Goryachev, Vladimir and Patramanis, Demosthenes and Krause, Michael and Redenti, Andrew and Aldous, David and Lai, Jesyin and Coleman, Shannon and Xu, Jiangnan and Lee, Sangwon and Magoulas, Ilias and Zhao, Sandy and Tang, Ning and Cohen, Michael K. and Paradise, Orr and Kirchner, Jan Hendrik and Ovchynnikov, Maksym and Matos, Jason O. and Shenoy, Adithya and Wang, Michael and Nie, Yuzhou and Sztyber-Betley, Anna and Faraboschi, Paolo and Riblet, Robin and Crozier, Jonathan and Halasyamani, Shiv and Verma, Shreyas and Joshi, Prashant and Meril, Eli and Ma, Ziqiao and Andréoletti, Jérémy and Singhal, Raghav and Platnick, Jacob and Nevirkovets, Volodymyr and Basler, Luke and Ivanov, Alexander and Khoury, Seri and Gustafsson, Nils and Piccardo, Marco and Mostaghimi, Hamid and Chen, Qijia and Singh, Virendra and Khánh, Tran Quoc and Rosu, Paul and Szlyk, Hannah and Brown, Zachary and Narayan, Himanshu and Menezes, Aline and Roberts, Jonathan and Alley, William and Sun, Kunyang and Patel, Arkil and Lamparth, Max and Reuel, Anka and Xin, Linwei and Xu, Hanmeng and Loader, Jacob and Martin, Freddie and Wang, Zixuan and Achilleos, Andrea and Preu, Thomas and Korbak, Tomek and Bosio, Ida and Kazemi, Fereshteh and Chen, Ziye and Bálint, Biró and Lo, Eve J. Y. and Wang, Jiaqi and Nunes, Maria Inês S. and Milbauer, Jeremiah and Bari, M. Saiful and Wang, Zihao and Ansarinejad, Behzad and Sun, Yewen and Durand, Stephane and Elgnainy, Hossam and Douville, Guillaume and Tordera, Daniel and Balabanian, George and Wolff, Hew and Kvistad, Lynna and Milliron, Hsiaoyun and Sakor, Ahmad and Eron, Murat and Andrew Favre, D. O. and Shah, Shailesh and Zhou, Xiaoxiang and Kamalov, Firuz and Abdoli, Sherwin and Santens, Tim and Barkan, Shaul and Tee, Allison and Zhang, Robin and Tomasiello, Alessandro and De Luca, G. Bruno and Looi, Shi-Zhuo and Le, Vinh-Kha and Kolt, Noam and Pan, Jiayi and Rodman, Emma and Drori, Jacob and Fossum, Carl J. and Muennighoff, Niklas and Jagota, Milind and Pradeep, Ronak and Fan, Honglu and Eicher, Jonathan and Chen, Michael and Thaman, Kushal and Merrill, William and Firsching, Moritz and Harris, Carter and Ciobâcă, Stefan and Gross, Jason and Pandey, Rohan and Gusev, Ilya and Jones, Adam and Agnihotri, Shashank and Zhelnov, Pavel and Mofayezi, Mohammadreza and Piperski, Alexander and Zhang, David K. and Dobarskyi, Kostiantyn and Leventov, Roman and Soroko, Ignat and Duersch, Joshua and Taamazyan, Vage and Ho, Andrew and Ma, Wenjie and Held, William and Xian, Ruicheng and Zebaze, Armel Randy and Mohamed, Mohanad and Leser, Julian Noah and Yuan, Michelle X. and Yacar, Laila and Lengler, Johannes and Olszewska, Katarzyna and Di Fratta, Claudio and Oliveira, Edson and Jackson, Joseph W. and Zou, Andy and Chidambaram, Muthu and Manik, Timothy and Haffenden, Hector and Stander, Dashiell and Dasouqi, Ali and Shen, Alexander and Golshani, Bita and Stap, David and Kretov, Egor and Uzhou, Mikalai and Zhidkovskaya, Alina Borisovna and Winter, Nick and Rodriguez, Miguel Orbegozo and Lauff, Robert and Wehr, Dustin and Tang, Colin and Hossain, Zaki and Phillips, Shaun and Samuele, Fortuna and Ekström, Fredrik and Hammon, Angela and Patel, Oam and Farhidi, Faraz and Medley, George and Mohammadzadeh, Forough and Peñaflor, Madellene and Kassahun, Haile and Friedrich, Alena and Perez, Rayner Hernandez and Pyda, Daniel and Sakal, Taom and Dhamane, Omkar and Mirabadi, Ali Khajegili and Hallman, Eric and Okutsu, Kenchi and Battaglia, Mike and Maghsoudimehrabani, Mohammad and Amit, Alon and Hulbert, Dave and Pereira, Roberto and Weber, Simon and {Handoko} and Peristyy, Anton and Malina, Stephen and Mehkary, Mustafa and Aly, Rami and Reidegeld, Frank and Dick, Anna-Katharina and Friday, Cary and Singh, Mukhwinder and Shapourian, Hassan and Kim, Wanyoung and Costa, Mariana and Gurdogan, Hubeyb and Kumar, Harsh and Ceconello, Chiara and Zhuang, Chao and Park, Haon and Carroll, Micah and Tawfeek, Andrew R. and Steinerberger, Stefan and Aggarwal, Daattavya and Kirchhof, Michael and Dai, Linjie and Kim, Evan and Ferret, Johan and Shah, Jainam and Wang, Yuzhou and Yan, Minghao and Burdzy, Krzysztof and Zhang, Lixin and Franca, Antonio and Pham, Diana T. and Loh, Kang Yong and Robinson, Joshua and Jackson, Abram and Giordano, Paolo and Petersen, Philipp and Cosma, Adrian and Colino, Jesus and White, Colin and Votava, Jacob and Vinnikov, Vladimir and Delaney, Ethan and Spelda, Petr and Stritecky, Vit and Shahid, Syed M. and Mourrat, Jean-Christophe and Vetoshkin, Lavr and Sponselee, Koen and Bacho, Renas and Yong, Zheng-Xin and De La Rosa, Florencia and Cho, Nathan and Li, Xiuyu and Malod, Guillaume and Weller, Orion and Albani, Guglielmo and Lang, Leon and Laurendeau, Julien and Kazakov, Dmitry and Adesanya, Fatimah and Portier, Julien and Hollom, Lawrence and Souza, Victor and Zhou, Yuchen Anna and Degorre, Julien and Yaln, Yiğit and Obikoya, Gbenga Daniel and Michael Pokorny, Rai and Bigi, Filippo and Boscá, M. C. and Shumar, Oleg and Bacho, Kaniuar and Recchia, Gabriel and Popescu, Mara and Shulga, Nikita and Tanwie, Ngefor Mildred and Lux, Thomas C. H. and Rank, Ben and Ni, Colin and Brooks, Matthew and Yakimchyk, Alesia and Quinn Liu, Huanxu and Cavalleri, Stefano and Häggström, Olle and Verkama, Emil and Newbould, Joshua and Gundlach, Hans and Brito-Santana, Leonor and Amaro, Brian and Vajipey, Vivek and Grover, Rynaa and Wang, Ting and Kratish, Yosi and Li, Wen-Ding and Gopi, Sivakanth and Caciolai, Andrea and De Witt, Christian Schroeder and Hernández-Cámara, Pablo and Rodolà, Emanuele and Robins, Jules and Williamson, Dominic and Raynor, Brad and Qi, Hao and Segev, Ben and Fan, Jingxuan and Martinson, Sarah and Wang, Erik Y. and Hausknecht, Kaylie and Brenner, Michael P. and Mao, Mao and Demian, Christoph and Kassani, Peyman and Zhang, Xinyu and Avagian, David and Scipio, Eshawn Jessica and Ragoler, Alon and Tan, Justin and Sims, Blake and Plecnik, Rebeka and Kirtland, Aaron and Bodur, Omer Faruk and Shinde, D. P. and Labrador, Yan Carlos Leyva and Adoul, Zahra and Zekry, Mohamed and Karakoc, Ali and Santos, Tania C. B. and Shamseldeen, Samir and Karim, Loukmane and Liakhovitskaia, Anna and Resman, Nate and Farina, Nicholas and Gonzalez, Juan Carlos and Maayan, Gabe and Anderson, Earth and De Oliveira Pena, Rodrigo and Kelley, Elizabeth and Mariji, Hodjat and Pouriamanesh, Rasoul and Wu, Wentao and Finocchio, Ross and Alarab, Ismail and Cole, Joshua and Ferreira, Danyelle and Johnson, Bryan and Safdari, Mohammad and Dai, Liangti and Arthornthurasuk, Siriphan and McAlister, Isaac C. and Moyano, Alejandro José and Pronin, Alexey and Fan, Jing and Ramirez-Trinidad, Angel and Malysheva, Yana and Pottmaier, Daphiny and Taheri, Omid and Stepanic, Stanley and Perry, Samuel and Askew, Luke and Rodrguez, Raúl Adrián Huerta and Minissi, Ali M. R. and Lorena, Ricardo and Iyer, Krishnamurthy and Fasiludeen, Arshad Anil and Clark, Ronald and Ducey, Josh and Piza, Matheus and Somrak, Maja and Vergo, Eric and Qin, Juehang and Borbás, Benjámin and Chu, Eric and Lindsey, Jack and Jallon, Antoine and McInnis, I. M. J. and Chen, Evan and Semler, Avi and Gloor, Luk and Shah, Tej and Carauleanu, Marc and Lauer, Pascal and Huy, Tran Duc and Shahrtash, Hossein and Duc, Emilien and Lewark, Lukas and Brown, Assaf and Albanie, Samuel and Weber, Brian and Vaz, Warren S. and Clavier, Pierre and Fan, Yiyang and Poesia Reis E Silva, Gabriel and Tony Lian, Long and Abramovitch, Marcus and Jiang, Xi and Mendoza, Sandra and Islam, Murat and Gonzalez, Juan and Mavroudis, Vasilios and Xu, Justin and Kumar, Pawan and Goswami, Laxman Prasad and Bugas, Daniel and Heydari, Nasser and Jeanplong, Ferenc and Jansen, Thorben and Pinto, Antonella and Apronti, Archimedes and Galal, Abdallah and Ze-An, Ng and Singh, Ankit and Jiang, Tong and Of Arc Xavier, Joan and Agarwal, Kanu Priya and Berkani, Mohammed and Zhang, Gang and Du, Zhehang and De Oliveira Junior, Benedito Alves and Malishev, Dmitry and Remy, Nicolas and Hartman, Taylor D. and Tarver, Tim and Mensah, Stephen and Loume, Gautier Abou and Morak, Wiktor and Habibi, Farzad and Hoback, Sarah and Cai, Will and Gimenez, Javier and Montecillo, Roselynn Grace and Łucki, Jakub and Campbell, Russell and Sharma, Asankhaya and Meer, Khalida and Gul, Shreen and Gonzalez, Daniel Espinosa and Alapont, Xavier and Hoover, Alex and Chhablani, Gunjan and Vargus, Freddie and Agarwal, Arunim and Jiang, Yibo and Patil, Deepakkumar and Outevsky, David and Scaria, Kevin Joseph and Maheshwari, Rajat and Dendane, Abdelkader and Shukla, Priti and Cartwright, Ashley and Bogdanov, Sergei and Mündler, Niels and Möller, Sören and Arnaboldi, Luca and Thaman, Kunvar and Siddiqi, Muhammad Rehan and Saxena, Prajvi and Gupta, Himanshu and Fruhauff, Tony and Sherman, Glen and Vincze, Mátyás and Usawasutsakorn, Siranut and Ler, Dylan and Radhakrishnan, Anil and Enyekwe, Innocent and Salauddin, Sk Md and Muzhen, Jiang and Maksapetyan, Aleksandr and Rossbach, Vivien and Harjadi, Chris and Bahaloohoreh, Mohsen and Sparrow, Claire and Sidhu, Jasdeep and Ali, Sam and Bian, Song and Lai, John and Singer, Eric and Uro, Justine Leon and Bateman, Greg and Sayed, Mohamed and Menshawy, Ahmed and Duclosel, Darling and Bezzi, Dario and Jain, Yashaswini and Aaron, Ashley and Tiryakioglu, Murat and Siddh, Sheeshram and Krenek, Keith and Shah, Imad Ali and Jin, Jun and Creighton, Scott and Peskoff, Denis and EL-Wasif, Zienab and P, Ragavendran and Richmond, Michael and McGowan, Joseph and Patwardhan, Tejal and Sun, Hao-Yu and Sun, Ting and Zubić, Nikola and Sala, Samuele and Ebert, Stephen and Kaddour, Jean and Schottdorf, Manuel and Wang, Dianzhuo and Petruzella, Gerol and Meiburg, Alex and Medved, Tilen and ElSheikh, Ali and Hebbar, S. Ashwin and Vaquero, Lorenzo and Yang, Xianjun and Poulos, Jason and Zouhar, Vilém and Bogdanik, Sergey and Zhang, Mingfang and Sanz-Ros, Jorge and Anugraha, David and Dai, Yinwei and Nhu, Anh N. and Wang, Xue and Demircali, Ali Anil and Jia, Zhibai and Zhou, Yuyin and Wu, Juncheng and He, Mike and Chandok, Nitin and Sinha, Aarush and Luo, Gaoxiang and Le, Long and Noyé, Mickaël and Perełkiewicz, Michał and Pantidis, Ioannis and Qi, Tianbo and Purohit, Soham Sachin and Parcalabescu, Letitia and Nguyen, Thai-Hoa and Winata, Genta Indra and Ponti, Edoardo M. and Li, Hanchen and Dhole, Kaustubh and Park, Jongee and Abbondanza, Dario and Wang, Yuanli and Nayak, Anupam and Caetano, Diogo M. and Wong, Antonio A. W. L. and Del Rio-Chanona, Maria and Kondor, Dániel and Francois, Pieter and Chalstrey, Ed and Zsambok, Jakob and Hoyer, Dan and Reddish, Jenny and Hauser, Jakob and Rodrigo-Ginés, Francisco-Javier and Datta, Suchandra and Shepherd, Maxwell and Kamphuis, Thom and Zhang, Qizheng and Kim, Hyunjun and Sun, Ruiji and Yao, Jianzhu and Dernoncourt, Franck and Krishna, Satyapriya and Rismanchian, Sina and Pu, Bonan and Pinto, Francesco and Wang, Yingheng and Shridhar, Kumar and Overholt, Kalon J. and Briia, Glib and Nguyen, Hieu and Quod Soler Bartomeu, David and Pang, Tony Cy and Wecker, Adam and Xiong, Yifan and Li, Fanfei and Huber, Lukas S. and Jaeger, Joshua and De Maddalena, Romano and Lù, Xing Han and Zhang, Yuhui and Beger, Claas and Kon, Patrick Tser Jern and Li, Sean and Sanker, Vivek and Yin, Ming and Liang, Yihao and Zhang, Xinlu and Agrawal, Ankit and Yifei, Li S. and Zhang, Zechen and Cai, Mu and Sonmez, Yasin and Cozianu, Costin and Li, Changhao and Slen, Alex and Yu, Shoubin and Park, Hyun Kyu and Sarti, Gabriele and Briański, Marcin and Stolfo, Alessandro and Nguyen, Truong An and Zhang, Mike and Perlitz, Yotam and Hernandez-Orallo, Jose and Li, Runjia and Shabani, Amin and Juefei-Xu, Felix and Dhingra, Shikhar and Zohar, Orr and Nguyen, My Chiffon and Pondaven, Alexander and Yilmaz, Abdurrahim and Zhao, Xuandong and Jin, Chuanyang and Jiang, Muyan and Todoran, Stefan and Han, Xinyao and Kreuer, Jules and Rabern, Brian and Plassart, Anna and Maggetti, Martino and Yap, Luther and Geirhos, Robert and Kean, Jonathon and Wang, Dingsu and Mollaei, Sina and Sun, Chenkai and Yin, Yifan and Wang, Shiqi and Li, Rui and Chang, Yaowen and Wei, Anjiang and Bizeul, Alice and Wang, Xiaohan and Arrais, Alexandre Oliveira and Mukherjee, Kushin and Chamorro-Padial, Jorge and Liu, Jiachen and Qu, Xingyu and Guan, Junyi and Bouyamourn, Adam and Wu, Shuyu and Plomecka, Martyna and Chen, Junda and Tang, Mengze and Deng, Jiaqi and Subramanian, Shreyas and Xi, Haocheng and Chen, Haoxuan and Zhang, Weizhi and Ren, Yinuo and Tu, Haoqin and Kim, Sejong and Chen, Yushun and Marjanović, Sara Vera and Ha, Junwoo and Luczyna, Grzegorz and Ma, Jeff J. and Shen, Zewen and Song, Dawn and Zhang, Cedegao E. and Wang, Zhun and Gendron, Gaël and Xiao, Yunze and Smucker, Leo and Weng, Erica and Lee, Kwok Hao and Ye, Zhe and Ermon, Stefano and Lopez-Miguel, Ignacio D. and Knights, Theo and Gitter, Anthony and Park, Namkyu and Wei, Boyi and Chen, Hongzheng and Pai, Kunal and Elkhanany, Ahmed and Lin, Han and Siedler, Philipp D. and Fang, Jichao and Mishra, Ritwik and Zsolnai-Fehér, Károly and Jiang, Xilin and Khan, Shadab and Yuan, Jun and Jain, Rishab Kumar and Lin, Xi and Peterson, Mike and Wang, Zhe and Malusare, Aditya and Tang, Maosen and Gupta, Isha and Fosin, Ivan and Kang, Timothy and Dworakowska, Barbara and Matsumoto, Kazuki and Zheng, Guangyao and Sewuster, Gerben and Villanueva, Jorge Pretel and Rannev, Ivan and Chernyavsky, Igor and Chen, Jiale and Banik, Deepayan and Racz, Ben and Dong, Wenchao and Wang, Jianxin and Bashmal, Laila and Gonçalves, Duarte V. and Hu, Wei and Bar, Kaushik and Bohdal, Ondrej and Patlan, Atharv Singh and Dhuliawala, Shehzaad and Geirhos, Caroline and Wist, Julien and Kansal, Yuval and Chen, Bingsen and Tire, Kutay and Yücel, Atak Talay and Christof, Brandon and Singla, Veerupaksh and Song, Zijian and Chen, Sanxing and Ge, Jiaxin and Ponkshe, Kaustubh and Park, Isaac and Shi, Tianneng and Ma, Martin Q. and Mak, Joshua and Lai, Sherwin and Moulin, Antoine and Cheng, Zhuo and Zhu, Zhanda and Zhang, Ziyi and Patil, Vaidehi and Jha, Ketan and Men, Qiutong and Wu, Jiaxuan and Zhang, Tianchi and Vieira, Bruno Hebling and Aji, Alham Fikri and Chung, Jae-Won and Mahfoud, Mohammed and Thi Hoang, Ha and Sperzel, Marc and Hao, Wei and Meding, Kristof and Xu, Sihan and Kostakos, Vassilis and Manini, Davide and Liu, Yueying and Toukmaji, Christopher and Yu, Eunmi and Demircali, Arif Engin and Sun, Zhiyi and Dewerpe, Ivan and Qin, Hongsen and Pflugfelder, Roman and Bailey, James and Morris, Johnathan and Heilala, Ville and Rosset, Sybille and Yu, Zishun and Chen, Peter E. and Yeo, Woongyeong and Jain, Eeshaan and Chigurupati, Sreekar and Chernyavsky, Julia and Reddy, Sai Prajwal and Venugopalan, Subhashini and Batra, Hunar and Park, Core Francisco and Tran, Hieu and Maximiano, Guilherme and Zhang, Genghan and Liang, Yizhuo and Shiyu, Hu and Xu, Rongwu and Pan, Rui and Suresh, Siddharth and Liu, Ziqi and Gulati, Samaksh and Zhang, Songyang and Turchin, Peter and Bartlett, Christopher W. and Scotese, Christopher R. and Cao, Phuong M. and Wu, Ben and Karwowski, Jacek and Scaramuzza, Davide},
	month = jan,
	year = {2026},
	pages = {1139--1146},
}

@article{paganelli_natural_2024,
	title = {Natural language processing in dermatology: {A} systematic literature review and state of the art},
	volume = {38},
	issn = {1468-3083},
	shorttitle = {Natural language processing in dermatology},
	url = {https://onlinelibrary.wiley.com/doi/abs/10.1111/jdv.20286},
	doi = {10.1111/jdv.20286},
	language = {en},
	number = {12},
	urldate = {2026-02-08},
	journal = {Journal of the European Academy of Dermatology and Venereology},
	author = {Paganelli, Alessia and Spadafora, Marco and Navarrete-Dechent, Cristian and Guida, Stefania and Pellacani, Giovanni and Longo, Caterina},
	year = {2024},
	note = {\_eprint: https://onlinelibrary.wiley.com/doi/pdf/10.1111/jdv.20286},
	pages = {2225--2234},
}

@misc{sellergren_medgemma_2025,
	title = {{MedGemma} {Technical} {Report}},
	url = {http://arxiv.org/abs/2507.05201},
	doi = {10.48550/arXiv.2507.05201},
	urldate = {2026-02-03},
	publisher = {arXiv},
	author = {Sellergren, Andrew and Kazemzadeh, Sahar and Jaroensri, Tiam and Kiraly, Atilla and Traverse, Madeleine and Kohlberger, Timo and Xu, Shawn and Jamil, Fayaz and Hughes, Cían and Lau, Charles and Chen, Justin and Mahvar, Fereshteh and Yatziv, Liron and Chen, Tiffany and Sterling, Bram and Baby, Stefanie Anna and Baby, Susanna Maria and Lai, Jeremy and Schmidgall, Samuel and Yang, Lu and Chen, Kejia and Bjornsson, Per and Reddy, Shashir and Brush, Ryan and Philbrick, Kenneth and Asiedu, Mercy and Mezerreg, Ines and Hu, Howard and Yang, Howard and Tiwari, Richa and Jansen, Sunny and Singh, Preeti and Liu, Yun and Azizi, Shekoofeh and Kamath, Aishwarya and Ferret, Johan and Pathak, Shreya and Vieillard, Nino and Merhej, Ramona and Perrin, Sarah and Matejovicova, Tatiana and Ramé, Alexandre and Riviere, Morgane and Rouillard, Louis and Mesnard, Thomas and Cideron, Geoffrey and Grill, Jean-bastien and Ramos, Sabela and Yvinec, Edouard and Casbon, Michelle and Buchatskaya, Elena and Alayrac, Jean-Baptiste and Lepikhin, Dmitry and Feinberg, Vlad and Borgeaud, Sebastian and Andreev, Alek and Hardin, Cassidy and Dadashi, Robert and Hussenot, Léonard and Joulin, Armand and Bachem, Olivier and Matias, Yossi and Chou, Katherine and Hassidim, Avinatan and Goel, Kavi and Farabet, Clement and Barral, Joelle and Warkentin, Tris and Shlens, Jonathon and Fleet, David and Cotruta, Victor and Sanseviero, Omar and Martins, Gus and Kirk, Phoebe and Rao, Anand and Shetty, Shravya and Steiner, David F. and Kirmizibayrak, Can and Pilgrim, Rory and Golden, Daniel and Yang, Lin},
	month = jul,
	year = {2025},
	note = {arXiv:2507.05201 [cs]},
}

@article{dymek_building_2021,
	title = {Building the evidence-base to reduce electronic health record–related clinician burden},
	volume = {28},
	copyright = {https://academic.oup.com/journals/pages/open\_access/funder\_policies/chorus/standard\_publication\_model},
	issn = {1527-974X},
	url = {https://academic.oup.com/jamia/article/28/5/1057/6042116},
	doi = {10.1093/jamia/ocaa238},
	language = {en},
	number = {5},
	urldate = {2026-02-03},
	journal = {Journal of the American Medical Informatics Association},
	author = {Dymek, Christine and Kim, Bryan and Melton, Genevieve B and Payne, Thomas H and Singh, Hardeep and Hsiao, Chun-Ju},
	month = apr,
	year = {2021},
	pages = {1057--1061},
}

@article{gandhi_how_2023,
	title = {How can artificial intelligence decrease cognitive and work burden for front line practitioners?},
	volume = {6},
	copyright = {https://creativecommons.org/licenses/by/4.0/},
	issn = {2574-2531},
	url = {https://academic.oup.com/jamiaopen/article/doi/10.1093/jamiaopen/ooad079/7255314},
	doi = {10.1093/jamiaopen/ooad079},
	language = {en},
	number = {3},
	urldate = {2026-02-03},
	journal = {JAMIA Open},
	author = {Gandhi, Tejal K and Classen, David and Sinsky, Christine A and Rhew, David C and Vande Garde, Nikki and Roberts, Andrew and Federico, Frank},
	month = jul,
	year = {2023},
	pages = {ooad079},
}

@misc{yang_qwen3_2025,
	title = {Qwen3 {Technical} {Report}},
	url = {http://arxiv.org/abs/2505.09388},
	doi = {10.48550/arXiv.2505.09388},
	urldate = {2026-02-02},
	publisher = {arXiv},
	author = {Yang, An and Li, Anfeng and Yang, Baosong and Zhang, Beichen and Hui, Binyuan and Zheng, Bo and Yu, Bowen and Gao, Chang and Huang, Chengen and Lv, Chenxu and Zheng, Chujie and Liu, Dayiheng and Zhou, Fan and Huang, Fei and Hu, Feng and Ge, Hao and Wei, Haoran and Lin, Huan and Tang, Jialong and Yang, Jian and Tu, Jianhong and Zhang, Jianwei and Yang, Jianxin and Yang, Jiaxi and Zhou, Jing and Zhou, Jingren and Lin, Junyang and Dang, Kai and Bao, Keqin and Yang, Kexin and Yu, Le and Deng, Lianghao and Li, Mei and Xue, Mingfeng and Li, Mingze and Zhang, Pei and Wang, Peng and Zhu, Qin and Men, Rui and Gao, Ruize and Liu, Shixuan and Luo, Shuang and Li, Tianhao and Tang, Tianyi and Yin, Wenbiao and Ren, Xingzhang and Wang, Xinyu and Zhang, Xinyu and Ren, Xuancheng and Fan, Yang and Su, Yang and Zhang, Yichang and Zhang, Yinger and Wan, Yu and Liu, Yuqiong and Wang, Zekun and Cui, Zeyu and Zhang, Zhenru and Zhou, Zhipeng and Qiu, Zihan},
	month = may,
	year = {2025},
	note = {arXiv:2505.09388 [cs]},
}

@article{obika_safety_2024,
	title = {Safety principles for medical summarization using generative {AI}},
	volume = {30},
	issn = {1546-170X},
	doi = {10.1038/s41591-024-03313-y},
	language = {eng},
	number = {12},
	journal = {Nature Medicine},
	author = {Obika, Dillon and Kelly, Christopher and Ding, Nicola and Farrance, Chris and Krause, Jonathan and Mittal, Praney and Cheung, Donny and Cole-Lewis, Heather and Elish, Madeleine and Karthikesalingam, Alan and Webster, Dale and Patel, Bakul and Howell, Michael},
	month = dec,
	year = {2024},
	pages = {3417--3419},
}

@article{bear_dont_walk_scoping_2022,
	title = {A scoping review of ethics considerations in clinical natural language processing},
	volume = {5},
	copyright = {https://creativecommons.org/licenses/by-nc/4.0/},
	issn = {2574-2531},
	url = {https://academic.oup.com/jamiaopen/article/doi/10.1093/jamiaopen/ooac039/6593700},
	doi = {10.1093/jamiaopen/ooac039},
	language = {en},
	number = {2},
	urldate = {2026-02-02},
	journal = {JAMIA Open},
	author = {Bear Don’t Walk, Oliver J and Reyes Nieva, Harry and Lee, Sandra Soo-Jin and Elhadad, Noémie},
	month = apr,
	year = {2022},
	pages = {ooac039},
}

@article{wang_systematic_2020,
	title = {Systematic {Evaluation} of {Research} {Progress} on {Natural} {Language} {Processing} in {Medicine} {Over} the {Past} 20 {Years}: {Bibliometric} {Study} on {PubMed}},
	volume = {22},
	shorttitle = {Systematic {Evaluation} of {Research} {Progress} on {Natural} {Language} {Processing} in {Medicine} {Over} the {Past} 20 {Years}},
	url = {https://www.jmir.org/2020/1/e16816},
	doi = {10.2196/16816},
	language = {EN},
	number = {1},
	urldate = {2026-02-02},
	journal = {Journal of Medical Internet Research},
	publisher = {JMIR Publications Inc., Toronto, Canada},
	author = {Wang, Jing and Deng, Huan and Liu, Bangtao and Hu, Anbin and Liang, Jun and Fan, Lingye and Zheng, Xu and Wang, Tong and Lei, Jianbo},
	month = jan,
	year = {2020},
	note = {Company: Journal of Medical Internet Research
Distributor: Journal of Medical Internet Research
Institution: Journal of Medical Internet Research
Label: Journal of Medical Internet Research},
	pages = {e16816},
}

@article{marwaha_algorithmic_2025,
	title = {The algorithmic consultant: a new era of clinical {AI} calls for a new workforce of physician-algorithm specialists},
	volume = {8},
	copyright = {2025 The Author(s)},
	issn = {2398-6352},
	shorttitle = {The algorithmic consultant},
	url = {https://www.nature.com/articles/s41746-025-01960-0},
	doi = {10.1038/s41746-025-01960-0},
	language = {en},
	number = {1},
	urldate = {2026-02-02},
	journal = {npj Digital Medicine},
	publisher = {Nature Publishing Group},
	author = {Marwaha, Jayson S. and Yuan, William and Poddar, Mukund and Elsamadisi, Pansy and Brat, Gabriel A.},
	month = aug,
	year = {2025},
	pages = {552},
}

@article{goodman-meza_natural_2022,
	title = {Natural {Language} {Processing} and {Machine} {Learning} to {Identify} {People} {Who} {Inject} {Drugs} in {Electronic} {Health} {Records}},
	volume = {9},
	issn = {2328-8957},
	url = {https://doi.org/10.1093/ofid/ofac471},
	doi = {10.1093/ofid/ofac471},
	number = {9},
	urldate = {2026-02-02},
	journal = {Open Forum Infectious Diseases},
	author = {Goodman-Meza, David and Tang, Amber and Aryanfar, Babak and Vazquez, Sergio and Gordon, Adam J and Goto, Michihiko and Goetz, Matthew Bidwell and Shoptaw, Steven and Bui, Alex A T},
	month = sep,
	year = {2022},
	pages = {ofac471},
}

@article{bootsma-robroeks_ai-generated_2025,
	title = {{AI}-generated draft replies to patient messages: exploring effects of implementation},
	volume = {7},
	issn = {2673-253X},
	shorttitle = {{AI}-generated draft replies to patient messages},
	url = {https://www.frontiersin.org/journals/digital-health/articles/10.3389/fdgth.2025.1588143/full},
	doi = {10.3389/fdgth.2025.1588143},
	language = {English},
	urldate = {2026-02-02},
	journal = {Frontiers in Digital Health},
	publisher = {Frontiers},
	author = {Bootsma-Robroeks, Charlotte M. H. H. T. and Workum, Jessica D. and Schuit, Stephanie C. E. and Hoekman, Anne and Mehri, Tarannom and Doornberg, Job N. and van der Laan, Tom P. and Schoonbeek, Rosanne C.},
	month = jun,
	year = {2025},
}

@misc{yilmaz_resource-efficient_2025,
	title = {Resource-efficient medical vision language model for dermatology via a synthetic data generation framework},
	copyright = {© 2025, Posted by Cold Spring Harbor Laboratory. This pre-print is available under a Creative Commons License (Attribution-NonCommercial 4.0 International), CC BY-NC 4.0, as described at http://creativecommons.org/licenses/by-nc/4.0/},
	url = {https://www.medrxiv.org/content/10.1101/2025.05.17.25327785v2},
	doi = {10.1101/2025.05.17.25327785},
	language = {en},
	urldate = {2026-02-02},
	publisher = {medRxiv},
	author = {Yilmaz, Abdurrahim and Yuceyalcin, Furkan and Varol, Rahmetullah and Gokyayla, Ece and Erdem, Ozan and Choi, Donghee and Demircali, Ali Anil and Gencoglan, Gulsum and Posma, Joram M. and Temelkuran, Burak},
	month = jul,
	year = {2025},
	note = {ISSN: 3067-2007
Pages: 2025.05.17.25327785},
}

@article{azarfar_responsible_2025,
	title = {Responsible adoption of multimodal artificial intelligence in health care: promises and challenges},
	volume = {7},
	issn = {2589-7500},
	shorttitle = {Responsible adoption of multimodal artificial intelligence in health care},
	url = {https://www.thelancet.com/journals/landig/article/PIIS2589-7500(25)00099-8/fulltext},
	doi = {10.1016/j.landig.2025.100917},
	language = {English},
	number = {12},
	urldate = {2026-02-02},
	journal = {The Lancet Digital Health},
	publisher = {Elsevier},
	author = {Azarfar, Ghazal and Naimimohasses, Sara and Rambhatla, Sirisha and Komorowski, Matthieu and Ferro, Diana and Lewis, Peter R. and Gates, Darren and Shara, Nawar and Gascon, Gregg M. and Chang, Anthony and Mamdani, Muhammad and Bhat, Mamatha},
	month = dec,
	year = {2025},
}

@article{goldberg_primer_2016,
	title = {A {Primer} on {Neural} {Network} {Models} for {Natural} {Language} {Processing}},
	volume = {57},
	copyright = {Copyright (c)},
	issn = {1076-9757},
	url = {https://www.jair.org/index.php/jair/article/view/11030},
	doi = {10.1613/jair.4992},
	language = {en},
	urldate = {2026-02-02},
	journal = {Journal of Artificial Intelligence Research},
	author = {Goldberg, Yoav},
	month = nov,
	year = {2016},
	pages = {345--420},
}

@article{schmidt_pemphigus_2019,
	title = {Pemphigus},
	volume = {394},
	issn = {0140-6736, 1474-547X},
	url = {https://www.thelancet.com/journals/lancet/article/PIIS0140-6736(19)31778-7/fulltext},
	doi = {10.1016/S0140-6736(19)31778-7},
	language = {English},
	number = {10201},
	urldate = {2026-02-02},
	journal = {The Lancet},
	publisher = {Elsevier},
	author = {Schmidt, Enno and Kasperkiewicz, Michael and Joly, Pascal},
	month = sep,
	year = {2019},
	pages = {882--894},
}

@article{wu_survey_2022,
	title = {A survey on clinical natural language processing in the {United} {Kingdom} from 2007 to 2022},
	volume = {5},
	issn = {2398-6352},
	url = {https://www.nature.com/articles/s41746-022-00730-6},
	doi = {10.1038/s41746-022-00730-6},
	language = {en},
	number = {1},
	urldate = {2026-02-02},
	journal = {npj Digital Medicine},
	author = {Wu, Honghan and Wang, Minhong and Wu, Jinge and Francis, Farah and Chang, Yun-Hsuan and Shavick, Alex and Dong, Hang and Poon, Michael T. C. and Fitzpatrick, Natalie and Levine, Adam P. and Slater, Luke T. and Handy, Alex and Karwath, Andreas and Gkoutos, Georgios V. and Chelala, Claude and Shah, Anoop Dinesh and Stewart, Robert and Collier, Nigel and Alex, Beatrice and Whiteley, William and Sudlow, Cathie and Roberts, Angus and Dobson, Richard J. B.},
	month = dec,
	year = {2022},
	pages = {186},
}

@article{van_veen_adapted_2024,
	title = {Adapted large language models can outperform medical experts in clinical text summarization},
	volume = {30},
	issn = {1078-8956, 1546-170X},
	url = {https://www.nature.com/articles/s41591-024-02855-5},
	doi = {10.1038/s41591-024-02855-5},
	language = {en},
	number = {4},
	urldate = {2026-02-02},
	journal = {Nature Medicine},
	author = {Van Veen, Dave and Van Uden, Cara and Blankemeier, Louis and Delbrouck, Jean-Benoit and Aali, Asad and Bluethgen, Christian and Pareek, Anuj and Polacin, Malgorzata and Reis, Eduardo Pontes and Seehofnerová, Anna and Rohatgi, Nidhi and Hosamani, Poonam and Collins, William and Ahuja, Neera and Langlotz, Curtis P. and Hom, Jason and Gatidis, Sergios and Pauly, John and Chaudhari, Akshay S.},
	month = apr,
	year = {2024},
	pages = {1134--1142},
}

@article{kasperkiewicz_pemphigus_2017,
	title = {Pemphigus},
	volume = {3},
	copyright = {2017 Macmillan Publishers Limited},
	issn = {2056-676X},
	url = {https://www.nature.com/articles/nrdp201726},
	doi = {10.1038/nrdp.2017.26},
	language = {en},
	number = {1},
	urldate = {2026-02-02},
	journal = {Nature Reviews Disease Primers},
	publisher = {Nature Publishing Group},
	author = {Kasperkiewicz, Michael and Ellebrecht, Christoph T. and Takahashi, Hayato and Yamagami, Jun and Zillikens, Detlef and Payne, Aimee S. and Amagai, Masayuki},
	month = may,
	year = {2017},
	pages = {17026},
}

\newpage

\section*{Appendix}
The main prompt and clinical features with their input type and notes. Each clinical feature were prompted separately. \\

\textbf{Main Prompt:}
You are an expert dermatologist tasked with summarizing long-term clinical notes for patients with pemphigus. Based on the provided input type, options, and notes (a, b, c, d, e), please complete the request by generating a horizontal CSV format that incorporates data from all patient visits. \\

\textbf{Clinical Features:}
\begin{enumerate}[label=\arabic*., leftmargin=*, itemsep=0.35em]

  \item \texttt{report\_id}, Open text

  \item \texttt{patient\_id}, Open text (unique patient identifier)

  \item \texttt{report\_span\_start\_date}, Date (YYYY-MM-DD) \,|\, NA

  \item \texttt{report\_span\_end\_date}, Date (YYYY-MM-DD) \,|\, NA

  \item \texttt{visit\_count\_est}, Integer \,|\, NA

  \item \texttt{pemphigus\_diagnosis}, Yes \,|\, No \,|\, NA

  \item \texttt{pemphigus\_subtype}, Pemphigus Vulgaris \,|\, Pemphigus Foliaceus \,|\, Paraneoplastic Pemphigus (PNP) \,|\, IgA pemphigus \,|\, Pemphigus Herpetiformis \,|\, Pemphigus Vegetans \,|\, NA

  \item \texttt{diagnosis\_confidence}, High \,|\, Low
    \begin{enumerate}[label=\alph*., leftmargin=2em, itemsep=0.2em]
  \item 1) ``High'' if at least 2 of the following are concordant: histology, direct immunofluorescence, anti-DSG1/3 values; 2) ``Low'' if 1 or none are concordant.
\end{enumerate}

\item \texttt{phenotype\_mucosal}, Yes \,|\, No \,|\, NA

\item \texttt{phenotype\_cutaneous}, Yes \,|\, No \,|\, NA

\item \texttt{current\_disease\_activity}, Active \,|\, Partial remission on treatment \,|\, Partial remission off treatment \,|\, Minimal treatment \,|\, Complete remission on treatment \,|\, Complete remission off treatment \,|\, NA
\begin{enumerate}[label=\alph*., leftmargin=2em, itemsep=0.2em]
  \item You need to decide \texttt{current\_disease\_activity} based on that information:
    \begin{enumerate}[label=\roman*., leftmargin=2.5em, itemsep=0.15em]
      \item Complete remission off therapy: No new lesions for at least 2 months without receiving any treatment.
      \item Complete remission on therapy: No new lesions while receiving minimal therapy.
      \item Minimal therapy: Prednisone 10 mg/day (or equivalent) for at least 2 months and/or minimal adjuvant therapy.
      \item Partial remission off therapy: Development of lesions that heal spontaneously within 1 week, occurring over a 2-month period without treatment.
      \item Partial remission on therapy: Development of lesions that heal within 1 week while receiving minimal therapy.
      \item Active (relapse/flare): While the disease is under control, the appearance of 3 or more new lesions per month that do not heal spontaneously within 1 week.
    \end{enumerate}
\end{enumerate}

\item \texttt{last\_flare\_date}, Date (YYYY-MM-DD) \,|\, No \,|\, NA

\item \texttt{flare\_count\_est}, Integer \,|\, NA
\begin{enumerate}[label=\alph*., leftmargin=2em, itemsep=0.2em]
  \item The initial presentation of the disease should not be counted, unless the patient received the initial diagnosis elsewhere and then came here.
\end{enumerate}

\item \texttt{histology\_pemphigus\_compatible}, Yes \,|\, No \,|\, NA

\item \texttt{dif\_pemphigus\_compatible}, Yes \,|\, No \,|\, NA

\item \texttt{anti\_dsg1\_positive\_at\_first\_diagnosis}, Numeric \,|\, Positive \,|\, Negative \,|\, NA

\item \texttt{anti\_dsg3\_positive\_at\_first\_diagnosis}, Numeric \,|\, Positive \,|\, Negative \,|\, NA

\item \texttt{anti\_dsg1\_last\_value}, Numeric \,|\, Positive \,|\, Negative \,|\, NA

\item \texttt{anti\_dsg1\_last\_date}, Date (YYYY-MM-DD) \,|\, NA

\item \texttt{anti\_dsg3\_last\_value}, Numeric \,|\, Positive \,|\, Negative \,|\, NA

\item \texttt{anti\_dsg3\_last\_date}, Date (YYYY-MM-DD) \,|\, NA

\item \texttt{current\_systemic\_treatment}, None \,|\, Steroid \,|\, Azathioprine \,|\, MMF \,|\, MTX \,|\, Cyclophosphamide \,|\, Rituximab \,|\, Combination \,|\, NA

\item \texttt{current\_prednisone\_mg}, Numeric \,|\, NA
\begin{enumerate}[label=\alph*., leftmargin=2em, itemsep=0.2em]
  \item If patient doesn’t have a treatment, mention 0.
\end{enumerate}

\item \texttt{current\_steroid\_duration\_months}, Numeric \,|\, NA

\item \texttt{total\_steroid\_duration\_months}, Numeric \,|\, NA

\item \texttt{azathioprine\_ever}, Yes \,|\, No

\item \texttt{mmf\_ever}, Yes \,|\, No

\item \texttt{methotrexate\_ever}, Yes \,|\, No

\item \texttt{cyclophosphamide\_ever}, Yes \,|\, No

\item \texttt{immunsupresant\_stop\_reason}, Ineffective \,|\, Adverse event \,|\, Remission \,|\, Other \,|\, NA

\item \texttt{rituximab\_ever}, Yes \,|\, No

\item \texttt{rituximab\_cycles\_count}, Integer

\item \texttt{rituximab\_first\_date}, Date (YYYY-MM-DD) \,|\, NA

\item \texttt{rituximab\_last\_date}, Date (YYYY-MM-DD) \,|\, NA

\item \texttt{rituximab\_response}, Complete \,|\, Partial \,|\, None \,|\, NA

\item \texttt{ivig\_ever}, Yes \,|\, No

\item \texttt{ivig\_last\_date}, Date (YYYY-MM-DD) \,|\, NA

\item \texttt{plasmapheresis\_ever}, Yes \,|\, No

\item \texttt{plasmapheresis\_last\_date}, Date (YYYY-MM-DD) \,|\, NA

\item \texttt{serious\_infection\_ever}, Yes \,|\, No \,|\, NA
\begin{enumerate}[label=\alph*., leftmargin=2em, itemsep=0.2em]
  \item Infections requiring hospitalization and intravenous treatment (treatment through a vein).
\end{enumerate}

\item \texttt{cytopenia\_ever}, Yes \,|\, No \,|\, NA
\begin{enumerate}[label=\alph*., leftmargin=2em, itemsep=0.2em]
  \item Hemogram (CBC): If available and there is no toxicity, write ``No''; if there is no data, write ``NA''.
\end{enumerate}

\item \texttt{liver\_toxicity\_ever}, Yes \,|\, No \,|\, NA
\begin{enumerate}[label=\alph*., leftmargin=2em, itemsep=0.2em]
  \item Biochemistry: If available and there is no toxicity, write ``No''; if there is no data, write ``NA''.
  \item Hepatosteatosis (fatty liver) does not indicate liver toxicity.
\end{enumerate}

\item \texttt{renal\_toxicity\_ever}, Yes \,|\, No \,|\, NA
\begin{enumerate}[label=\alph*., leftmargin=2em, itemsep=0.2em]
  \item Biochemistry: If available and there is no toxicity, write ``No''; if there is no data, write ``NA''.
\end{enumerate}

\item \texttt{infusion\_reaction\_ever}, Yes \,|\, No \,|\, NA

\item \texttt{steroid\_complications\_ever}, Yes \,|\, No \,|\, NA

\item \texttt{steroid\_complication\_hyperglycemia}, Yes \,|\, No \,|\, NA
\begin{enumerate}[label=\alph*., leftmargin=2em, itemsep=0.2em]
  \item If there is suspicion of steroid complications, you can mark the criterion as ``Yes''.
\end{enumerate}

\item \texttt{steroid\_complication\_hypertension}, Yes \,|\, No \,|\, NA
\begin{enumerate}[label=\alph*., leftmargin=2em, itemsep=0.2em]
  \item If there is suspicion of steroid complications, you can mark the criterion as ``Yes''.
\end{enumerate}

\item \texttt{steroid\_complication\_osteoporosis\_or\_osteopenia}, Osteoporosis \,|\, Osteopenia \,|\, No \,|\, NA
\begin{enumerate}[label=\alph*., leftmargin=2em, itemsep=0.2em]
  \item If there is suspicion of steroid complications, you can mark the criterion as ``Yes''.
  \item T-score: below $-1$ indicates osteopenia; below $-2.5$ indicates osteoporosis.
\end{enumerate}

\item \texttt{steroid\_complication\_cataract}, Yes \,|\, No \,|\, NA
\begin{enumerate}[label=\alph*., leftmargin=2em, itemsep=0.2em]
  \item If there is suspicion of steroid complications, you can mark the criterion as ``Yes''.
\end{enumerate}

\item \texttt{steroid\_complication\_glaucoma}, Yes \,|\, No \,|\, NA
\begin{enumerate}[label=\alph*., leftmargin=2em, itemsep=0.2em]
  \item If there is suspicion of steroid complications, you can mark the criterion as ``Yes''.
\end{enumerate}

\item \texttt{steroid\_complication\_myopathy}, Yes \,|\, No \,|\, NA
\begin{enumerate}[label=\alph*., leftmargin=2em, itemsep=0.2em]
  \item If there is suspicion of steroid complications, you can mark the criterion as ``Yes''.
\end{enumerate}

\item \texttt{steroid\_complication\_avascular\_necrosis}, Yes \,|\, No \,|\, NA
\begin{enumerate}[label=\alph*., leftmargin=2em, itemsep=0.2em]
  \item If there is suspicion of steroid complications, you can mark the criterion as ``Yes''.
\end{enumerate}

\item \texttt{medication\_start\_stop\_dates\_present}, Yes \,|\, Partial \,|\, No \,|\, NA

\item \texttt{dose\_changes\_documented}, Yes \,|\, No \,|\, NA

\item \texttt{longitudinal\_course\_clear}, Yes \,|\, No \,|\, NA

\item \texttt{current\_status}, Open text
\begin{enumerate}[label=\alph*., leftmargin=2em, itemsep=0.2em]
  \item You need to write a plain text report based on that information (one single paragraph, no bullet list). Answer: pemphigus type; current disease activity; last eye exam date; last bone density test date; osteoporosis treatment at visit (if valid); last Quantiferon date; INH prophylaxis date (if given); last hepatitis serology date; last CBC date; last LFT date; rituximab ever (if yes, last rituximab date); current prednisone use; any safety/adverse events; treatments received and dates; side effects (immunosuppressant-related infections; steroid complications: hyperglycemia, hypertension, osteoporosis, cataract, glaucoma, myopathy, avascular necrosis, other; rituximab complications: infusion reaction, cytopenia, serious infection, other). Also include major surgery (cardiac surgery, organ transplantation, etc.) and comorbid conditions. If there is suspicion of steroid complications, mention it in the summary.
  \item If repeated entries occur but the date stays the same while text changes (suggesting multiple admissions), warn: (1) if both date and text are the same $\rightarrow$ duplicate entry; (2) if date is the same but text is different $\rightarrow$ possible date error.
  \item The treatment summary should indicate the start year and end year of treatment. If unclear, state that it is unclear.
\end{enumerate}

\end{enumerate}

\end{document}